# A Uniqueness Theorem for Clustering

**Reza Bosagh Zadeh**
School of Computer Science
Carnegie Mellon University,
Pittsburgh, Pennsylvania, USA
rezab@cs.cmu.edu

**Shai Ben-David**
David R. Cheriton School of Computer Science
University of Waterloo,
Waterloo, Ontario, Canada
shai@cs.uwaterloo.ca

## Abstract

Despite the widespread use of Clustering, there is distressingly little general theory of clustering available. Questions like "What distinguishes a clustering of data from other data partitioning?", "Are there any principles governing all clustering paradigms?", "How should a user choose an appropriate clustering algorithm for a particular task?", etc. are almost completely unanswered by the existing body of clustering literature. We consider an axiomatic approach to the theory of Clustering. We adopt the framework of Kleinberg, [Kle03]. By relaxing one of Kleinberg's clustering axioms, we sidestep his impossibility result and arrive at a consistent set of axioms. We suggest to extend these axioms, aiming to provide an axiomatic taxonomy of clustering paradigms. Such a taxonomy should provide users some guidance concerning the choice of the appropriate clustering paradigm for a given task. The main result of this paper is a set of abstract properties that characterize the Single-Linkage clustering function. This characterization result provides new insight into the properties of desired data groupings that make Single-Linkage the appropriate choice. We conclude by considering a taxonomy of clustering functions based on abstract properties that each satisfies.

## 1 Introduction

The task of clustering is a basic data analysis technique in many fields. Whenever one has a set of objects with an underlying measure of similarity, it is natural to seek a grouping of these objects in groups that respect these similarities. While the clustering task thus described arises in a very wide range of different applications, it is, of course, ill defined. Given a specific data set there are often many different ways to partition that data in a way that adheres to the above description. Consequently, there is an overwhelming variety of clustering paradigms and techniques currently in use.

The majority of work in the study of clustering has so far been in the context of concrete methods, concrete data generative models or concrete data-sets. However, there is very little research considering clustering from a more general perspective, research into unifying principles that are common to all clustering paradigms as well as research comparing different clustering paradigms in terms of their characterizing features. Our research aims to address these higher level aspects of clustering.

We wish to reason about clustering independently of any particular algorithm, objective function, or generative data model. We define a clustering function as one that satisfies a set of properties. This is often termed as an *axiomatic* framework. Functions that satisfy the basic axioms required to be called a *clustering function* can then further classified according to other properties to provide its user with guidance concerning which such function best suits their application.

Kleinberg [Kle03] sets forth three properties that one may expect all clustering functions to satisfy, but then proves no function can satisfy all three properties. By shifting the formalism of that discussion to clustering quality measures, [ABD08] concludes that Kleinberg's impossibility result is more an artifact of the formalism and framework in which the result was stated, rather than a natural and inherent contradiction in our intuition about clustering. In the current paper, we circumvent Kleinberg's impossibility result by relaxing one of his axioms. We then propose some additional abstract properties of clustering functions and obtain a uniqueness theorem, showing that only the Single-Linkage function satisfies the resulting set of properties. Our result highlights the aspects of a similarity measure over a data set that Single-Linkage is focused on.

We adopt the framework and the axioms proposed by [Kle03], however, we circumvent the impossibility result by restricting our attention to clustering algorithms that take the number of clusters to be created as part of their input. It is known that such a restriction suffices to render the resulting set of axioms consistent. In fact, most of the common clustering algorithms, such as Single-Linkage and Min-Sum $k$-clustering, satisfy this relaxed set of axioms.

It is clear that fixing the number of clusters beforehand restricts the set of functions that can be described by the framework. However, one interpretation of Kleinberg's impossibility result is that if one does not give algorithms the number of clusters they are to return, then the algorithm must be performing some unintuitive operations. Furthermore, many algorithms used in practice require the number of clusters. Indeed, *all* objective functions ($k$-means, $k$-median, ...) used in practice require $k$, along with many hierarchical (Single-Linkage, Complete Linkage, ...) and spectral algorithms.

Our characterization of the Single-Linkage (SL) clustering is in terms of two other notions:

- Given a weighted graph, a *minimum spanning tree* is a weighted tree which touches every node, while minimizing the sum of the edges used in the tree. [GR69] showed that all the information needed to compute the Single-Linkage $k$-partitioning of a graph is present in the minimum spanning tree (MST) of the graph. Given a weighted graph and $k$, the Single-Linkage algorithm returns a $k$-partitioning on the nodes. It is easy to see that Single-Linkage clustering of a data set can be obtained by computing the MST of the induced complete graph (where edges are weighted by the distance between their end points), then cutting the $k-1$ most expensive edges of the MST; leaving behind exactly $k$ disconnected components, which are returned as clusters. It follows, that if two distance functions give rise to the same MST then Single-Linkage will return the same clustering on both. We'll prove that Single-Linkage is *the only* clustering function that has this property, on top of satisfying some other natural clustering axioms.

- Given a dissimilarity measure, $d$ over some domain set $X$, we define the *d-induced path distance*, $P_d$, by setting, for all $x, y \in X$, $P_d(x, y) = \min_{q \in P_{x,y}} \max_{i < |q|} d(q(i), q(i+1))$ (where $|q|$ denotes the number of vertices in a path $q$, and $q(i)$ is the $i$'th vertex in the path). We say that a clustering function, $F$, is *path-distance-coherent* if, for any pair of distance measures $d, d'$ over the same domain, if $d$ and $d'$ induce the same path distance (namely, for all $x, y$, it is the case that $P_d(x, y) = P_{d'}(x, y)$) then, for all $k$, $F(d, k) = F(d', k)$.

It is easy to see that Single-Linkage is a path-distance-coherent clustering function. Furthermore, it is not hard to realize that if $d$ and $d'$ give rise to the same MST then they induce the same path distance (namely, for all $x, y$, $P_d(x, y) = P_{d'}(x, y)$). It follows that if a clustering function is path-distance coherent it is also MST -coherent. Our characterization of Single-Linkage in terms of MST therefore implies a similar characterization in terms of path coherence. Namely, Single-Linkage is *the only* clustering function that is path-distance-coherent, on top of satisfying some other natural clustering axioms.

Path distance coherence can be viewed as a natural property that users may wish their choice clustering function to satisfy. For example, consider the case a data set of objects that have been subject to some noisy perturbations, allowing an object to be replaced by another if their distance is small. This may be the case in a data set of DNA sequences of different species under some edit distance, or a set of folklore documents that have been copied over and over again. In such cases, a user may require that a clustering groups together objects that might have a common ancestor, and can be formalized as objects with a small path-distance between them. Our message to the user is that in such a case, Single-Linkage should be their choice clustering function.

The outline of this paper is as follows: First we lay down formal foundations for our properties. After defining the properties, to investigate the ramifications of exhibiting algorithms which satisfy a subset of these properties, in section 2 we introduce the Minimum Spanning Tree Cuts family of partitioning functions. Then we prove the uniqueness theorem (theorem 5), investigate the properties satisfied by the Min-Sum $k$-clustering objective in section 3, and conclude with a proposed taxonomy of partitioning functions (table 1).

**More related work**

As mentioned above, our formal framework is based on Kleinberg's [Kle03]. We also adopt two of the three axioms proposed in that paper, Consistency and Scale Invariance. We replace the Richness paper of [Kle03] by its version for the case of fixed numebr of clusters, $k$-Richness.

Another related axiomatic approach is the work of Jardine and Sibson [JS75]. They constrain their view of clustering functions and only consider hierarchical functions. They show that Single-Linkage is the only function satisfying a set of properties; however, this is primarily a consequence of the fact that one of their properties is an implicit optimization criterion that is uniquely optimized by Single-Linkage. Our collection of properties are drastically different from [JS75] and do not restrict the formalism to only consider hierarchical functions. Furthermore, none of our properties are implicitly optimizing the Single-Linkage criterion, and we show that important subsets of our properties are not enough to characterize Single-Linkage; meaning all 4 properties are necessary to achieve uniqueness.

**Formal Preliminaries**

A partitioning function acts on a set $S$ of $n \geq 2$ points along with an integer $k > 0$, and pairwise distances among the points in $S$. The points in $S$ are not assumed to belong to any specific set; the pairwise distances are the only data the partitioning function has about them. Since we wish to deal with point sets that do not necessarily belong to a specific set, we identify the points with the set $S = \{1, 2, ..., n\}$. We can then define a distance function to be any function $d : S \times S \to R$ such that for distinct $i, j \in S$, we have $d(i, j) \geq 0, d(i, j) = 0$ if and only if $i = j$, and $d(i, j) = d(j, i)$ - in other words $d$ must be symmetric and two points have distance zero if only if they are the same point.

Sometimes we write $d = \langle e_1, e_2, \ldots, e_{\binom{n}{2}} \rangle$ to mean the set of edges that exist between all pairs of $n$ points. This list is *always* ordered by increasing weight. $w(e)$ is the weight of edge $e$ which connects some two points $i, j$. So $w(e) = d(i, j)$.

A partitioning function is a function $F$ that takes a distance function $d$ on $S \times S$ and an integer $k \geq 1$ and returns a $k$-partitioning of $S$. A $k$-partitioning of $S$ is a collection of non-empty disjoint subsets of $S$ whose union is $S$. The sets in $F(d, k)$ will be called its *clusters*. Two clustering functions are equivalent if and only if they output the same partitioning on all values of $d$ and $k$ - i.e. functionally equivalent. Now we describe the Single-Linkage partitioning function.

It should be noted that the behavior of Single-Linkage is robust against small fluctuations in the weight of the edges in $d$, so long as the order of edges does not change. This can be seen readily from algorithm 1.

Finally, an edge $e$ of $d$ is called **redundant** if the two ends of $e$ (call them $x, y$) are connected via a path whose edges are all individually of smaller weight than

**Algorithm 1** SINGLE-LINKAGE$(d, k)$
   Input: $d = \langle e_1, e_2, \ldots, e_{\binom{n}{2}} \rangle$, $k \geq 1$.
   Output: The Single-Linkage $k$-partitioning.

   $\Gamma \leftarrow \{\{1\}, \{2\}, \ldots, \{n\}\}$
4: $i \leftarrow 1$
   **while** $|\Gamma| > k$ **do**
       let $x, y$ be the two ends of $e_i$
       let $c_x \in \Gamma$, $c_y \in \Gamma$ be the clusters of $x$ and $y$
8:     **if** $c_x \neq c_y$ **then**
           Merge $c_x$ and $c_y$
           $\Gamma \leftarrow (\Gamma \backslash c_x, c_y) \cup \{c_x \cup c_y\}$
       **end if**
12:    $i \leftarrow i + 1$
   **end while**
   Output $\Gamma$

$w(e)$. These are exactly those edges which have $c_x = c_y$, and thus fail the line 8 condition of algorithm 1.

## 2 Uniqueness Theorem

Now in an effort to distinguish clustering functions from partitioning functions, we lay down some properties that one may like a clustering function to satisfy. Here is the first one. If $d$ is a distance function, then we define $\alpha \cdot d$ to be the same function with all distances multiplied by $\alpha$.

SCALE-INVARIANCE. *For any distance function $d$, number of clusters $k$, and scalar $\alpha > 0$, we have*
$$F(d, k) = F(\alpha \cdot d, k)$$

This property simply requires the function to be immune to stretching or shrinking the data points linearly. It effectively disallows clustering functions to be sensitive to changes in units of measurement - which is desirable. We would like clustering functions to not have any predefined hard-coded distance values in their decision process. This property was initially used in an axiomatic clustering framework by [JS75].

The next property is one that is exhibited by algorithms such as {Single, Complete, Average}-linkage. Let the order of edges in a distance function $d$ be defined as the ordering induced by the weights of the edges.

ORDER-CONSISTENCY. *For any two distance functions $d$ and $d'$, number of clusters $k$, if the order of edges in $d$ is the same as the order of edges in $d'$, then $F(d, k) = F(d', k)$*

This essentially means that the only way that the function uses edge weights is by comparing them against each other. Functions that do arithmetic on the edge weights (e.g. the $k$-median objective function) often do not satisfy this property, whereas algorithms such as Complete-Linkage do. Notice that Order-Consistency implies Scale-Invariance - meaning that any function which satisfies Order-Consistency will also satisfy Scale-Invariance. For this reason, whenever a function satisfies Order-Consistency as well as Scale-Invariance, we don't mention Scale-Invariance explicitly. This property was initially considered by [JS75].

The next property ensures that the clustering function is "rich" in types of partitioning it could output. For a fixed $S$ and $k$, Let $\text{Range}(F(\bullet, k))$ be the set of all possible outputs while varying $d$.

$k$-RICHESS. *For any number of clusters $k$, $\text{Range}(F(\bullet, k))$ is equal to the set of all $k$-partitions of $S$*

In other words, if we are given a set of points such that all we know about the points are pairwise distances, then for any partitioning $\Gamma$, there should exist a $d$ such that $F(d, k) = \Gamma$. By varying distances amongst points, we should be able to obtain all possible $k$-partitionings.

The next property is more subtle and was initially introduced in [Kle03], along with richness. We call a partitioning function "consistent" if it satisfies the following: when we shrink distances between points in the same cluster and expand distances between between points in different clusters, we get the same result. Formally, we say that $d'$ is a $\Gamma$-*transformation* of $d$ if (a) for all $i, j \in S$ belonging to the same cluster of $\Gamma$, we have $d'(i, j) \leq d(i, j)$; and (b) for all $i, j \in S$ belonging to different clusters of $\Gamma$, we have $d'(i, j) \geq d(i, j)$. In other words, $d'$ is a transformation of $d$ such that points inside the same cluster are brought closer together and points not inside the same cluster are moved further away from one another.

CONSISTENCY. *Fix $k$. Let $d$ be a distance function, and $d'$ be a $F(d, k)$-transformation of $d$. Then $F(d, k) = F(d', k)$*

In other words, suppose that we run the partitioning function $F$ on $d$ to get back a particular partitioning $\Gamma$. Now, with respect to $\Gamma$, if we shrink in-cluster distances or expand between-cluster distances and run $F$ again, we should still get back the same result - namely $\Gamma$.

The difference between these and the properties defined in [Kle03] is that at all times, the partitioning function $F$ is forced to return a fixed number of clusters: $k$. If this were not the case, then the above 4 properties could never be satisfied by any function. In most popular clustering algorithms such as $k$-means, Single-Linkage, and spectral clustering, the number of clusters to be returned is determined beforehand – by the human user or other methods – and passed into the clustering function as a parameter. Kleinberg mentions this fact, but since the goal of [Kle03] was an impossibility result, $k$ was not provided to partitioning functions, thereby making the properties unsatisfiable.

Proving characterization of an object by some axioms always comes in two parts: satisfiability and uniqueness. Since satisfiability is suspect as a result of [Kle03], we now prove that our properties are simultaneously satisfiable, even though a simple modification of the framework would render them unsatisfiable. An overview of clustering functions and the properties each satisfies is given in table 1.

**Theorem 1.** *Single-Linkage is Consistent, $k$-Rich, Scale-Invariant, and Order-Consistent.*

*Proof.* Single-Linkage is Order-Consistent because it performs all its decisions by simply comparing two edges to see which is smaller/larger. Under Order-Consistent changes to the input, the decisions made by SL do not change, and therefore neither will its output. Scale-Invariance follows immediately since Order-Consistency is satisfied.

For $k$-Richness, when aiming to obtain a $k$-partitioning $\Gamma$, it is enough to set $\Gamma$ in-cluster distances to 1, and between cluster distances to 2 - then SL will return $\Gamma$. What remains is to show Consistency.

To show show consistency for SL, let $SL(d, k) = \Gamma$. With respect to $\Gamma$, we call edges with each side in two different clusters *outer* edges, and edges with both ends inside a cluster *inner* edges. Consider that to construct $\Gamma$, SL sorts all the edges of the graph, and successively looks at each edge: while there are more than $k$ clusters, SL turns the smallest outer edge into an inner edge (therefore reducing the number of clusters by 1). An inner edge that is larger than any outer edge is called a *redundant inner edge* because during the Single-Linkage process, the edge is not considered for merging, but nevertheless becomes an inner edge as a result of transitivity. Let the edges of the input graph be represented as $E = \langle e_1, e_2, \ldots, e_{\binom{n}{2}} \rangle$, in ascending order. Each of these edges can either be an outer edge, a non-redundant inner edge, or a redundant inner edge. By the SL process, we know that there exists prefix of $E$ (call it $P$) which will be all inner edges, and will be enough to define $\Gamma$. If $k = n$, then $P$ will be empty as there are no inner edges. Now consider all possible $\Gamma$-transformations of $d$. If we shrink a non-redundant inner edge of $d$, then $P$ will not change and the SL process will still produce $\Gamma$. If we shrink a redundant

|  | Scale-Invariance | Consistency | $k$-Richness | MST-Coherence | Order-Consistency |
|---|:---:|:---:|:---:|:---:|:---:|
| Single-Linkage | ✓ | ✓ | ✓ | ✓ | ✓ |
| MST cuts family | ✓ | × | ✓ | ✓ | ✓ |
| Min-Sum $k$-clustering | ✓ | ✓ | ✓ | × | × |
| Constant partitioning | ✓ | ✓ | × | ✓ | ✓ |

**Table 1:** Overview of discussed partitioning functions. Even if one were to consider more partitioning functions, as a consequence of theorem 5, the only row which can have all entries ✓ is Single-Linkage.

inner edge, then $P$ may change and become $P'$, but the clustering produced will not change as a result of transitivity. Finally, if we expand an outer edge of $E$, then again $P$ will not change thereby leaving $\Gamma$ intact. So in all possible $\Gamma$-transformations of $d$, we have $\text{SL}(d, k) = \text{SL}(d', k)$. [1] □

**MST-Coherence and the MST Cuts family**

As we observed from theorem 1, Consistency, Scale Invariance, and $k$-Richness do not uniquely determine a Single-Linkage since at least SL and MSKC – which are different clustering functions – satisfy all 3 properties. Two clustering functions are equivalent if and only if they output the same partitioning on all values of $d$ and $k$. To uniquely characterize SL, we introduce a new property. For ease of notation, let $\text{MST}(d)$ be the minimum spanning tree of the graph induced by $d$. $\text{MST}(d)$ is defined as the output of Kruskal's algorithm, with some some arbitrary tie-breaking strategy that is universal to all mentions of MST.

MST-COHERENCE. *If $d$ and $d'$ are distance functions such that MST(d) = MST(d'), then for all $k$, $F(d, k) = F(d', k)$*

In other words, if two datasets have the same minimum spanning tree, then for a partitioning function $F$ to be MST-Coherent, it must be that $F$ returns the same partitioning on both datasets. Note that for two minimum spanning trees to be equal, we must have all nodes, edges, and edge weights involved be equal and of the same weight. MST-Coherence is effectively forcing algorithms to ignore redundant edges, as is apparent in step 6 of the proof of theorem 5. MST-Coherence is not a property that we expect of all clustering functions, so we will never refer to it as an axiom. It is important to see that the properties are simultaneously satisfiable.

**Theorem 2.** *Single-Linkage satisfies MST-Coherence, Consistency, $k$-Richness, and Order-Consistency.*

---
[1] Kleinberg [Kle03] treated SL as a Consistent function without exhibiting a proof; we include a proof here for completeness.

*Proof.* Consistency was proven in theorem 1. $k$-Richness, MST-Coherence, and Order-Consistency are proven as part of a more general result in the proof of theorem 3. □

We will prove the main uniqueness theorem shortly. Before doing this, we reflect on the relationships between subsets of these properties and show that in addition to Single-Linkage, there exist other partitioning functions satisfying three of the four properties. But the only function that can satisfy all four is Single-Linkage. For the purpose of theorem 3, a property X is 'necessary' if all remaining properties together (3 of the 4) are not enough to characterize Single-Linkage.

**Theorem 3.** *Consistency, and $k$-Richness are necessary to characterize Single-Linkage.*

*Proof.* For each of the mentioned properties, we show that all the other properties are not enough to uniquely characterize Single-Linkage. For each of the properties, we exhibit an algorithm that acts differently than Single-Linkage, and satisfies all the properties except for one. In other words, we show that without each of these properties, the remaining ones do not uniquely characterize Single-Linkage.

We describe the **Minimum Spanning Tree Cuts** (MSTC) procedure, which in fact defines a family of clustering functions. As usual, the task is to partition $n$ points into $k$ clusters. Consider the family of clustering functions produced by computing the minimum spanning tree of the graph, and then cutting away $k-1$ predefined edges in set $C_k$. These $k-1$ edges are defined by the ascending order of their length in the minimum spanning tree. For example, if the set of edges to be cut were defined by $C_k = \{1, 2, 3, \ldots, k-1\}$, $|C_k| = k-1$, the algorithm would cut the shortest edge of the MST, then the second shortest edge, the third, all the way until the $(k-1)$'th edge has been cut, leaving exactly $k$ components, each representing a cluster. As another example, setting $C_k = \{n-1, n-2, n-3, \ldots, n-k+1\}$ defines Single-Linkage, since SL is exactly the algorithm where we compute the MST then cut away the $k-1$ most expensive edges. The entire family of MSTC functions are obtained by varying $C_k$. As we will see,

Single-Linkage is the only Consistent member of the MSTC family.

*Consistency is necessary.* We claim that the entire MSTC family of partitioning functions are all MST-Coherent, $k$-Rich, and Order-Consistent. However, it is not true that all members of MSTC are consistent - the only consistent function in MSTC is Single-Linkage. Thus Consistency is necessary to characterize SL.

We prove $k$-Richness is satisfied because we can design the edges we need to be cut to be exactly the edges in $C_k$: consider a target partitioning $\Gamma$. We may construct a $d$ such that all distances are very large, except for exactly one spanning tree $T$. Within $T$, we could like to cut any edge $e$ that is not an in-cluster edge with respect to $\Gamma$. We may do so by picking the size of $e$ such that its position with respect to the other edges in the MST is inside $C_k$, thus guaranteeing $e$ will be cut by the particular MSTC function. Once such a $d$ has been constructed, we can see that indeed $F(d) = \Gamma$.

MST-Coherence follows as a result of the fact the decisions made by an MSTC function are entirely dependent on the minimum spanning tree of the input - so if the MST doesn't change, then neither will the output. Order-Consistency is satisfied because the order and identity of the edges in the MST do not change, even though their absolute weights might, and cutting elements of $C_k$ is done by referring to the position of the edges in the edge-ordering. However, theorem 5 tells us that the only Consistent function within the MSTC family is Single-Linkage. As an example of a member of MSTC that is not Consistent, take $C_k = \{1, 2, 3, \ldots, k-1\}$.

*$k$-Richness is necessary.* Now consider the Constant clustering function which always returns the first $n - k + 1$ elements of $S$ as a single cluster and returns the remaining $k - 1$ as singleton clusters with one point inside each cluster, making a total of $k$ clusters. Because this function does not look at $d$, it is trivially MST-Coherent, Consistent, and Order-Consistent. However, it is not $k$-Rich because it always returns some singletons - i.e. we could never reach a $k$-partitioning that has no singletons (a singleton is a cluster with a single in it). $\square$

Now that we have seen our properties do not trivially characterize Single-Linkage, we can move onto proving the uniqueness theorem.

**Lemma 4.** *Given a Consistent partitioning function $F$, and a distance function $d$ with edges in ascending order of weight*

$$d = \langle e_1, e_2, \ldots, e_p, e_q, \ldots e_{\binom{n}{2}} \rangle$$

*then for all $k > 0$, if $e_p$ and $e_q$ are both inner edges or both outer edges (w.r.t. $F(d, k)$), we have*

$$F(\langle e_1, e_2, \ldots, e_q, e_p, \ldots e_{\binom{n}{2}} \rangle, k) = F(d, k)$$

*Proof.* In other words, whenever we have two edges of the same type (inner or outer), in neighboring positions in the edge ordering of $d$, we can swap their positions while maintaining the output of $F$. This is true because if both $e_p$ and $e_q$ are outer edges, then we can expand $e_p$ until $w(e_p) > w(e_q)$ all the while preserving the output of $F$ (by Consistency). Similarly, if both $e_p$ and $e_q$ are inner edges, we can shrink $e_q$ until $w(e_p) > w(e_q)$. $\square$

**Theorem 5.** *Single-Linkage is the only Consistent, $k$-Rich, MST-Coherent, Order-Consistent partitioning function.*

*Proof.* Let $F$ be any Consistent, $k$-Rich, MST-Coherent partitioning function, and let $d$ be any distance function on $n$ points. We want to show that that for all $k > 0$, $F(d, k) = \mathrm{SL}(d, k)$. For this purpose, we introduce the partitioning $\Gamma$ as whatever the output of SL is on $d$ and $k$, so $\mathrm{SL}(d, k) = \Gamma$. Whenever we say "inner" or "outer" edge for this proof, we mean with respect to $\Gamma$.

By $k$-Richness of $F$, there exists a $d_1$ such that $F(d_1, k) = \mathrm{SL}(d, k) = \Gamma$. Now, through a series of transformations that preserve the output of $F$, we transform $d_1$ into $d_2$, then $d_2$ into $d_3$, …, until we arrive at $d$. Let $d_i$ be represented by an ordered list of its edges in ascending order $d_i = \langle e_1^i, e_2^i, \ldots, e_{\binom{n}{2}}^i \rangle$.

By the definition of Single-Linkage, we know that a prefix $p = \langle e_1, e_2, \ldots, e_t \rangle$ of $d$ are all inner edges, where $t$ is exactly how many edges were considered for merging by the SL algorithm until $k$ clusters were formed. Once the edges of $p$ have all been declared inner edges, then all other inner edges will follow by transitivity. Since all inner edges of $\Gamma$ are identified by $p$, then $p$ is enough to define $\Gamma$.

Recall that redundant edges were defined earlier. We define *redundant inner edges* to mean any inner edges that have weight greater than some outer edge. Such inner edges are not part of $p$. Redundant inner edges cannot be part of the minimum spanning tree for $d$, since during Kruskal's algorithm for MST, the two components that a redundant inner edge connects have already been connected by non-redundant and cheaper inner edges. Since redundant inner edges are not part of the MST of $d$, increasing their weights will maintain this property, and so by increasing the weight of a redundant inner edge we do not change the MST.

Since we didn't change the MST, the output of $F$ will be preserved, by MST-Coherence.

Now we begin the $\Gamma$-preserving transformations on $d_1$ to eventually transform $d_1$ into $d$ while at each step $i$ maintaining $F(d_i, k) = \Gamma$.

1. By $k$-Richness, we know there exists a $d_1$ such that $F(d_1, k) = \text{SL}(d, k) = \Gamma$.

2. Since all edges of $p$ are inner edges, we can shrink them in $d_1$ until their position in the edge ordering of $d_1$ is $\leq t$. Call this newly created dataset $d_2$. $d_2$ has the same first $t$ edges as $d$, though not necessarily in the same order. This step maintains $F(d_2, k) = \Gamma$ by Consistency (we only shrank inner edges).

3. Now we reorder the first $t$ edges of to be in the exactly the same order as they appear in $d$. Call the new dataset $d_3$. This step maintains $F(d_3, k) = \Gamma$ by lemma 4 (all the first $t$ edges are of the same type - namely inner edges, so we may reorder them freely). Now the first $t$ edges of $d_3$ are the same edges (their weights may differ, but their identities match). Now we deal with the remaining $\binom{n}{2} - t$ edges.

4. We expand all outer edges until they are larger than all inner edges and call the result $d_4$. This step maintains $F(d_4, k) = \Gamma$ by Consistency.

5. Now we reorder all outer edges until their order in relation to each other is as they appear in $d$, and call the result $d_5$. This step maintains $F(d_5, k) = \Gamma$ by lemma 4. Now, $d_5$ has all the non-redundant inner edges in the same position as they appear in $d$, and has all the outer edges in the same order relative to one another as they appear in $d$. The only edges that are out-of-place are the redundant inner edges. The redundant inner edges need to be put into their correct position amongst the outer edges.

6. The redundant inner edges need to have their weights increased by the exactly the right amount so that they end up in the correct position in $d$. This can done simply by expanding these edges. As we mentioned earlier, by MST-Coherence, expanding redundant inner edges maintains $\Gamma$. So we call the this new data set $d_6$, and note that by MST-Coherence $F(d_6, k) = \Gamma$.

7. At this point in the edge ordering of $d_6$, all the edges lie in the same position as they do in $d$. However, their weights might be different than what appears in $d$. By using Order-Consistency, we can turn the weights of $d_6$ into exactly those of $d$, and call the result $d_7$. Since we didn't change the order of edges from $d_6$, by Order-Consistency we have that $F(d_7, k) = \Gamma$. It should be clear that $d_7 = d$.

8. Thus we have $F(d_7, k) = F(d, k) = \Gamma$.

We started with any $d$ and $k$, and showed that

$$F(d, k) = \Gamma = \text{SL}(d, k)$$

Also, in theorem 1 we showed that Single-Linkage satisfies all 4 properties. Thus it is uniquely characterized. □

Note that in the current paper when we mention Spanning Trees, we include the weights of the edges involved in the spanning tree as part of the identity of the tree - meaning that for two spanning trees to be equal they must have all edge weights equal. If we were to relax this requirement, then MST-Coherence would imply Order-Consistency, but then even Single-Linkage would not satisfy MST-Coherence. By considering edge weights when evaluating equality of two spanning trees we weaken the restrictions imposed on the partitioning function; which is desirable because it expands the applicability of theorem 5.

## 3 Building a taxonomy of clustering functions

Now that we know there is a single function satisfying all 4 properties in table 1, it is natural to ask which clustering functions satisfy subsets of these properties. We consider one example, the Min-Sum $k$-clustering objective function. Variations on the Min-Sum $k$-clustering objective are widely used. For now we focus on the Min-Sum $k$-clustering objective function itself leave the variations for future work. The objective is to minimize

$$\Lambda_d(\Gamma) = \sum_{C \in \Gamma} \sum_{i,j \in C} d(i,j)$$

Which is equivalent to the balanced $k$-median and Graph Cuts objective functions [BCR01, vL07]. Now, we investigate which properties are satisfied by MSKC.

**Theorem 6.** *Min-Sum $k$-clustering is Consistent, $k$-Rich, and Scale-invariant, but is neither Order-Consistent nor MST-Coherent.*

*Proof.* Scale-Invariance follows as a result of the optimal argument of the objective function being invariant to linear scaling. For $k$-Richness, when aiming to obtain a $k$-partitioning $\Gamma$, it is enough to set $\Gamma$ in-cluster distances to 1, and between cluster distances to 2 - then MSKC will return $\Gamma$. What remains is to show Consistency.

To show consistency for MSKC, we observe that for a fixed $k$, the set of all $k$-partitionings is an anti-chain as defined in [Kle03]. Therefore by the construction in proof of theorem 3.2 in [Kle03], MSKC is Consistent. Now we show that MSKC is neither MST-Coherent nor Order-Consistent.

We prove this for $k = 2$, the case for $k > 2$ is similar. Consider the distance $d$ function on $n$ (even) points. We arbitrarily split the points into 2 parts each of size $n/2$, called $A$ and $B$. Now we will construct $d$. For all $x, y \in B$, $d(x, y) = \epsilon$ for some $0 < \epsilon < 1$. And for all points $x \in A$, $y \in B$, we have $d(x, y) = 2$. Finally, all points inside $x, y \in A$ have distance $d(x, y) = \epsilon$ except for a single edge, between $x_0 \in A$, and $y_0 \in A$, for which we have $d(x_0, y_0) = 3$. So it is clear that all edges are either of length $\epsilon$ or 2, and there is a single edge of length 3. We now use this $d$ to show Min-Sum $k$-clustering is neither Order-Consistent nor MST-Coherent.

For sufficiently large $n$, if we run Min-Sum on $d$ with $k = 2$, we get back as answer the partitioning $\Gamma = \{A, B\}$, since it is cheaper to group $x_0$ and $y_0$ together than to pay $n/2 \times 2$. Notice that the edge between $x_0$ and $y_0$ is the largest edge of $d$. Thus, increasing the weight of $(x_0, y_0)$ is both an MST-Coherent and Order-Consistent transformation of $d$. Now, if we expand $(x_0, y_0)$ until its weight is larger than $3n$, then Min-Sum $k$-clustering will put $x_0$ and $y_0$ into different clusters, thus outputting a partitioning different from $\Gamma$, which violates Order-Consistency and MST-Coherence. $\square$

## 4 Conclusions & future directions

In this paper we make a step towards the important and ambitious goal of developing a general theory of clustering. More concretely, we aim to build a suite of abstract properties of clustering that will induce a taxonomy of clustering paradigms. Such a taxonomy should serve to help utilize prior domain knowledge to allow educated choice of a clustering method that is appropriate for a given clustering task.

At this point, we present a consistent axiomatic basis for clustering, and, under that framework, we go on to offer a concrete set of properties that characterize the Single-Linkage clustering algorithm. Our contribution provides new insight into the connection between Minimum Spanning Trees and the Single-Linkage clustering functions. This uniqueness result is set in a framework similar to one used for the purpose of showing impossibility, showing that there is no inherent impossibility in formalizing clustering. By considering the listing in table 1, we demonstrate the type of desired taxonomy of clustering functions based on the properties each satisfies. To investigate the ramifications of algorithms which satisfy only a subset of our properties, we introduced the Minimum Spanning Tree Cuts family of partitioning functions, of which Single-Linkage is the only Consistent member. The uniqueness theorem came about as a result of forcing functions to ignore redundant edges in the course of their operation.

We hope that this work spurs new research in building a more solid theory of Clustering, especially an axiomatic theory. Future considerations that will be valuable are the addition of new clustering functions and properties to table 1. We have characterized Single-Linkage in a useful way, leaving the characterization of other clustering functions (such as Min-Sum $k$-clustering) for future work.

### Acknowledgements

Thanks to Avrim Blum, Margareta Ackerman, and Nathan Schneider for very valuable discussions. Supported by NSF Grant OCI-0838385, and a graduate fellowship from NSERC. Part of this work was completed while Reza was an undergraduate at the University of Waterloo.